# BERT Modeli ile COVID-19 Aşısı için Aşı Karşıtı Tweetlerin Arttığı Zaman Aralıklarının Tespiti

Ülkü Tuncer KÜÇÜKTAŞ [1*], Fatih UYSAL[2], Fırat HARDALAÇ [3], İsmail BİRİ [4]

[1] Elektrik-Elektronik Mühendisliği, Gazi Üniversitesi, Türkiye Cumhuriyeti, ulku.tuncer.kucuktas@gazi.edu.tr
[2] Elektrik-Elektronik Mühendisliği, Gazi Üniversitesi, Türkiye Cumhuriyeti, uysal@gazi.edu.tr
[3] Elektrik-Elektronik Mühendisliği, Gazi Üniversitesi, Türkiye Cumhuriyeti, firat@gazi.edu.tr
[4] Genel Cerrahi, Koru Ankara Hastanesi, Türkiye Cumhuriyeti, genelcer44@hotmail.com

**Özet**

Covid-19'a karşı sunulan çözümlerden en etkili olanı geliştirilen çeşitli aşılardır. Aşılara duyulan güvensizlik bu çözümün hızlı ve etkili kullanılmasına engel oluşturabilir. Toplumun düşüncelerini ifade etme araçlarından birinin sosyal medya olduğu söylenebilir. Sosyal medyada aşı karşıtlığının arttığı zaman aralıklarının belirlenmesi, kurumların aşı karşıtlığı ile mücadele ederken kullanacağı stratejiyi belirlemelerinde yardımcı olabilir. Ancak sosyal medya için girilen tüm tweetlerin takibinin ve kaydının insan gücü ile yapılması, çok iş gücü gerektireceğinden ve verimsiz olacağından dolayı çeşitli otomasyon çözümlerine ihtiyaç duyulmaktadır. Bu çalışma kapsamında derin öğrenme tabanlı bir doğal dil işleme (natural language processing) NLP modeli olan dönüştürücülerden çift yönlü kodlayıcı temsilleri (bidirectional encoder representations from transformers) BERT modeli kullanılmıştır. Haber, anlamsızlar, aşı karşıtlığı ve aşı destekleyenler olmak üzere 4 farklı kategoriye (sınıfa) ayrılmış 1506 tweetlik bir veri setinde model, 25 devir (epoch) boyunca 5e-6 eğitim oranıyla (learning rate) eğitilmiştir. Eğitilmiş model kullanılarak aşı karşıtı tweetlerin yoğunlaştığı aralıkların tespiti için 652840 tweetin ait olduğu kategoriler belirlenmiştir. Belirlenen kategorilerin zaman içerisinde değişimi görselleştirilip, değişime neden olabilecek olaylar belirlenmiştir. Model eğitimi sonucunda, test veri setinde 0.81 f-skoru ve sınıflar için ROC eğrisi altında kalan kalanlar (AUC değerleri) sırasıyla 0.99 / 0.91 / 0.92 / 0.92 olarak elde edilmiştir. Bu modelde, literatürdeki çalışmalardan farklı olarak bu tarz tweetlerin tespit edilip sansürlenmesinden farklı bir amaç ile belli bir zaman aralığında aşı karşıtı tweetlerin zamana göre atılma sıklığı ölçülüp görselleştirerek kurumların strateji belirlerken kullanabilecekleri veri sağlayan yardımcı bir sistem tasarlanmıştır.

*Anahtar Sözcükler: Covid-19; Derin Öğrenme; Doğal Dil İşleme; Twitter; Veri Madenciliği.*

**Abstract**

The most effective of the solutions against Covid-19 is the various vaccines developed. Distrust of vaccines can hinder the rapid and effective use of this remedy. One of the means of expressing the thoughts of society is social media. Determining the time intervals during which anti-vaccination increases in social media can help institutions determine the strategy to be used in combating anti-vaccination. Recording and tracking every tweet entered with human labor would be inefficient, so various automation solutions are needed. In this study, The Bidirectional Encoder Representations from Transformers (BERT) model, which is a deep learning-based natural language processing (NLP) model, was used. In a dataset of 1506 tweets divided into four different categories as news, irrelevant, anti-vaccine, and vaccine supporters, the model was trained with a learning rate of 5e-6 for 25 epochs. To determine the intervals in which anti-vaccine tweets are concentrated, the categories to which 652840 tweets belong were determined by using the trained model. The change of the determined categories overtime was visualized and the events that could cause the change were determined. As a result of model training, in the test dataset, the f-score of 0.81 and AUC values for different classes were obtained as 0.99/0.91/0.92/0.92, respectively. In this model, unlike the studies in the literature, an auxiliary system is designed that provides data that institutions can use when determining their strategy by measuring and visualizing the frequency of anti-vaccine tweets in a time interval, different from detecting and censoring such tweets.

*Keywords: Covid-19; Deep Learning; Natural Language Processing; Twitter; Data Mining.*

1. **Giriş ve Amaç**

Covid-19 dünya üzerinde milyonlarca ölüme sebebiyet vermesinin yanında devletleri sosyal ve ekonomik hayata zarar verecek önlemler almak zorunda bırakması ile insanlığa maddi ve manevi zararlar vermiştir. Covid-19 için sunulan çözümlerden en etkili olanlarına geliştirilen aşılara toplumun belirli bir kısmı şüphe duymaktadır. Literatürde islamofobi ve aşı karşıtlığı gibi zararlı görülebilecek konuların tespiti ve sansürü için tasarlanmış makine öğrenmesi algoritmaları bulunmaktadır. Bu çalışma kapsamında literatürde bulunan çalışmalardan farklı olarak aşı ile ilgili genel kanının zaman içerisinde nasıl değiştiğinin incelenebileceği bir araç geliştirilmeye çalışılmıştır. Literatürde aynı konuda tespit için kullanılan modellerden en iyi performans gösteren BERT olduğu için, bu çalışma kapsamında da BERT modeli kullanılmıştır [1].22 Temmuz 2021 ve 22 Ağustos 2021 tarihleri arasında atılmış tüm tweetler kazınarak, eğitilen BERT modeli ile kategorileri belirlenmiş ve zamana göre değişimleri görselleştirilmiştir.

**1.1 Literatür Özeti**

To ve arkadaşları, aşı karşıtı tweetlerin tespiti için BERT , Bi-LSTM (long short-term memory), destek vektör makineleri (support vector machine, SVM), ve saf bayes (navie bayes, NB) sınıflandırma metodlarını karşılaştırmışlardır. Toplam 1,651,687 tweet içeren çalışmanın sonucunda BERT 0.95 f-skoru ile en büyük başarıyı göstermiştir [1]. Ruiz ve arkadaşları tarafından, doğal dil işleme yöntemleri kullanılarak aşı karşıtı ve aşı destekleyen internet grupları ve bu grupların birbiri ile olan ilişkisi incelenmiştir [2]. Ayan ve arkadaşları tarafından, twitter üzerinde islamofobi içeren tweetlerin tespiti için yapılan çalışmada 162.000 tweet içeren bir veri seti ile NB ve Ridge Regresyonu kullanmıştır [3]. Xue ve arkadaşları, twitter üzerinde covid-19 ile alakalı 4 milyon tweet üzerinde, Latent Dirichlet Allocation (LDA) yaklaşımını kullanarak incelemelerde bulunmuştur [4]. Budhwani ve Sun, Başkan Trump'ın "Çin virüsü" tweetini attığı 16 Mart 2020'den önceki ve sonraki tweet tartışmalarını karşılaştırmış ve Amerika Birleşik Devletleri'ndeki birçok eyalette insanların tweetlerinde "Çin virüsü" kullanımının önemli ölçüde arttığını tespit etmiştir [5].

**2. Materyal ve Yöntem**

**2.1. Veri Seti**

Twitter üzerinde paylaşılan 22 Temmuz 2021 - 22 Ağustos 2021 tarihleri arasında, içerisinde aşı sözcüğü geçen toplam 652840 tweet , snscrape kütüphanesi kullanan bir Python scripti ile çekilmiştir. Snscrape, sosyal ağ hizmetleri için bir kazıyıcıdır (scraper). Kullanıcı profilleri, hashtag'ler veya aramalar gibi şeyleri kazır ve kazılan öğeleri döndürür. Veri madenciliği için kullanılan araçlardan bir tanesidir [6]. Daha sonra bu kazınan tweetlerin içerisinden 1506 tanesi Doğal Dil İşleme modelinin eğitimi için rastgele seçilip 4 kategoriye ayrılmıştır. Veri setinin %80'i eğitim aşamasında, %20'si ise test aşamasında kullanılmıştır. Aşağıda belirtilen Tablo 1'de her bir kategoriye ilişkin olarak açıklamalar, miktarlar ve hastag'ler yer almaktadır.

| Kategoriler | Açıklama | Adet | Popüler Hastag'ler |
|---|---|---|---|
| Haber | Kurumların yaptığı açıklamaları ve istatistikleri içeren tweetler | 93 | #asiol #sondakika #haber |
| Anlamsızlar | Aşı kelimesi içeren ancak fikir belirtmeyen tweetler | 406 | |
| Aşı Karşıtları | Aşı karşıtı tweetler | 583 | #asiolmayacağım #asizorbalığınadiren |
| Aşı Destekleyenler | Aşı olunması yönünde tavsiye veren tweetler | 424 | #asiol #asiolalim #birliktebasaracagiz |

**Tablo 1**

**2.2. Model**

Aşı karşıtı tweetlerin tespiti için derin öğrenme temelli bir doğal dil işleme modeli kullanılmıştır. Bu model sınıflandırma ve soru cevaplama gibi geniş yelpazede görevler için kullanılabilen BERT modelidir. BERT modeli sadece son bir ek katmanını eğitilerek, kullanılabilir. Modelimiz için, MDZ Dijital Kütüphane ekibi (dbmdz) tarafından sağlanan önceden eğitilmiş bir tokenizer ve önceden eğitilmiş bir Türkçe BERT modeli kullanılmıştır [7]. BERT modeli diğer modellerden farklı olarak cümleyi hem sağdan sola hem soldan sağa değerlendirmeye çalışır ve bu sayede daha iyi sonuçlar elde etmeyi amaçlar. Bert modelinin ön-eğitiminde (pre-train) NSP (next sentence prediction) denilen bir yöntem, Wikipedia ve BookCorpus gibi veri setleri kullanılır [8] .

Çalışma kapsamında kullanılan veri seti ile, BERT modelinde ince ayar (fine-tune) yapılarak, 25 devir (epoch) boyunca 5e-6 öğrenme oranı (learning-rate) ile ADAM optimizer kullanılarak eğitilmiştir. Eğitim tamamlandıktan sonra model kullanarak 652840 tweet için kategoriler belirlenmiştir ve belirlenen kategoriler Python matplotlib kütüphanesi ile görselleştirilmiştir.

**3. Bulgular**

Veri setinde bulunan tweetlerin zaman göre atılma sıklığı, modelin 652840 tweet için yaptığı tahminler kullanılarak elde edilen kategorilerin zaman içinde değişimi ve aşı karşıtı tweetlerin zaman içerisinde yüzde olarak değişimi Şekil 1'de bulunabilir.

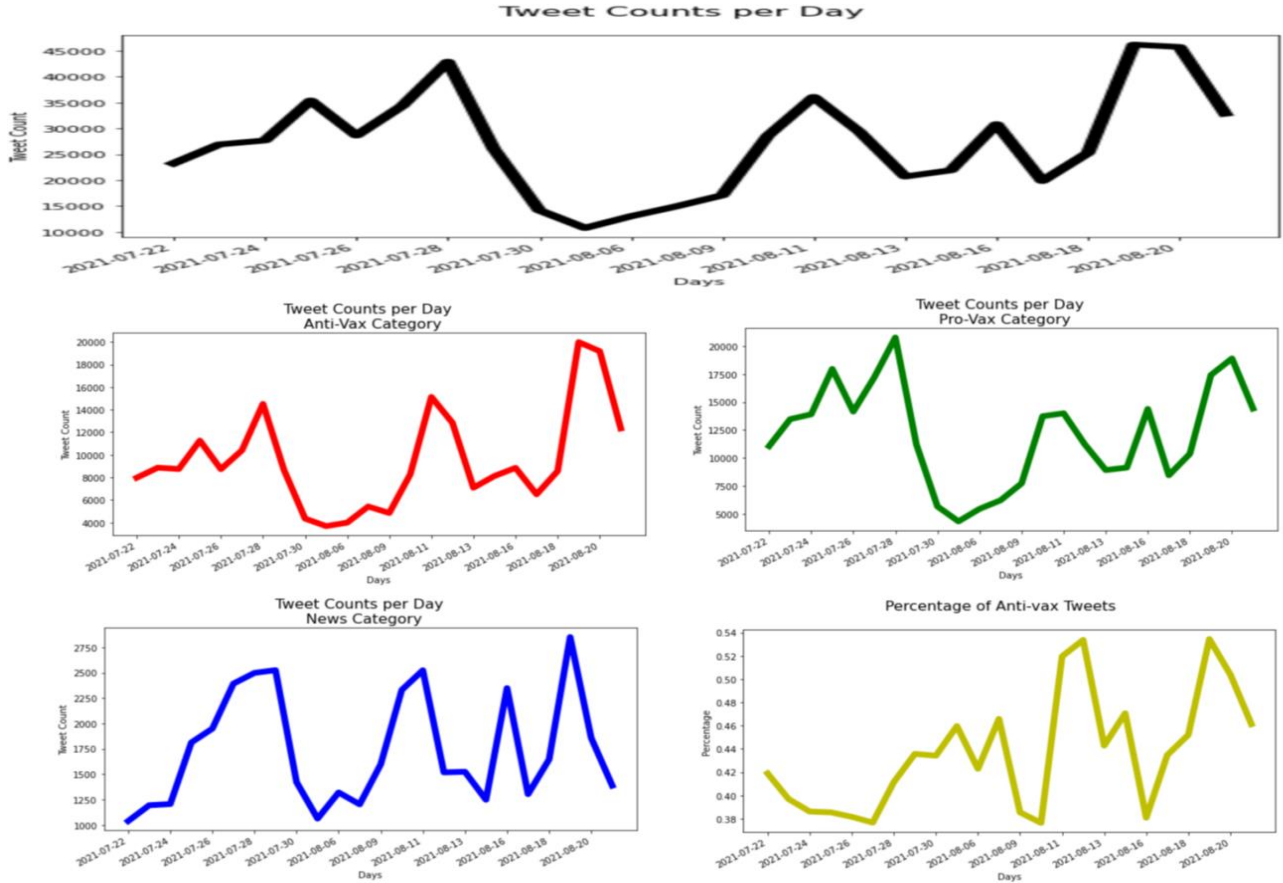

**Şekil 1**

Modelin test veri seti ile değerlendirilmesi sonucu elde edilen ROC-eğrileri ve "Confusion matrix" i Şekil 2'de gösterilmiştir

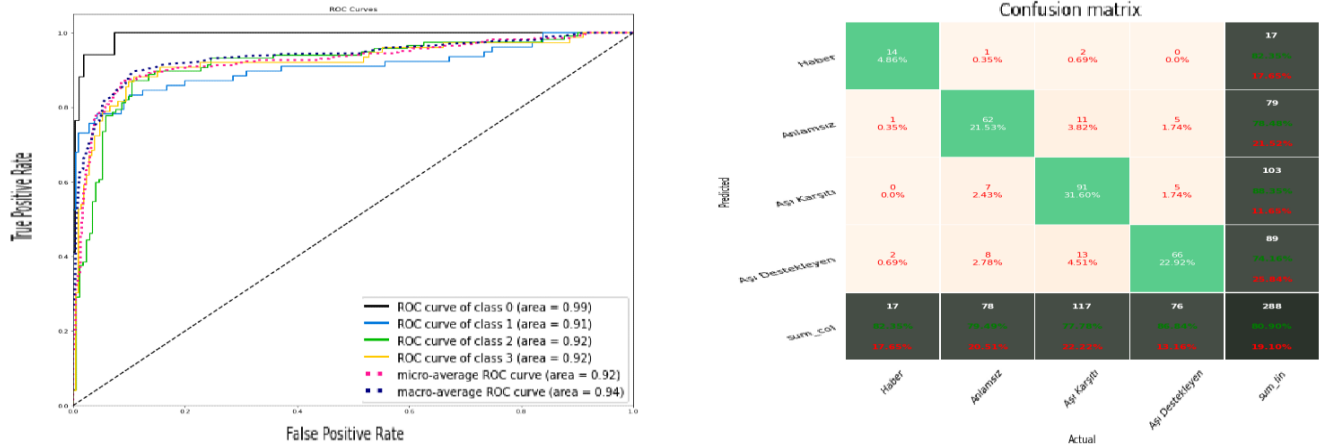

Şekil 2

Modelin sınıflara göre f1-skoru, recall ve precision değerleri Şekil 3'de gösterilmiştir.

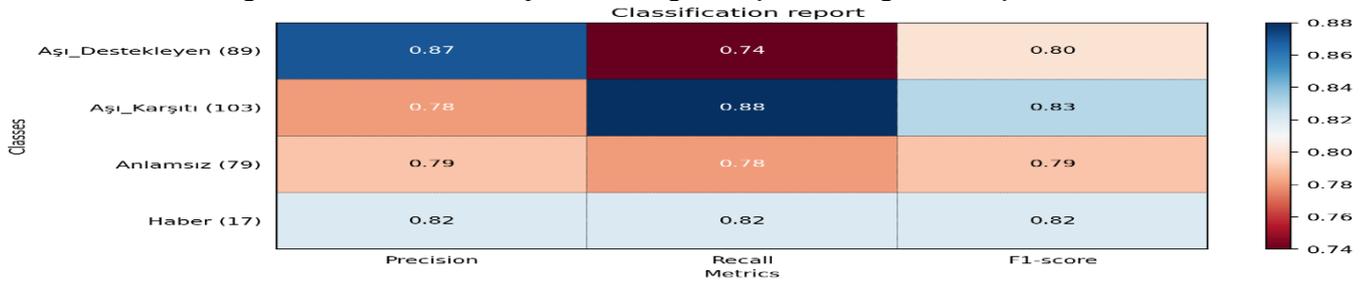

Şekil 3

## 4. Sonuçlar ve Öneriler

BERT dil işleme modeli, 1506 tweetten oluşan veri seti ile eğitiminin sonucunda tweetleri kategorilere ayırabilmiştir. Daha sonra 652840 tweetin kategorileri model ile belirlenmiş ve zamana göre değişimleri görselleştirilmiştir. Aşı olmayı destekleyen tweetlerin yüzde olarak en yüksek olduğu tarih 28 Temmuz ve Aşı karşıtı tweetlerin yüzde olarak en yüksek olduğu zamanlar 11 ve 19 Ağustos olarak tespit edilmiştir.
1) T.C. Sağlık Bakanı Sayın Dr. Fahrettin KOCA'nın 28 Temmuz 2021 tarihli Açıklaması [9]
2) T.C. Cumhurbaşkanı Sayın Recep Tayyip ERDOĞAN'ın 11 ve 19 Ağustos 2021 tarihli Açıklamaları [10,11]

Aşı ile ilgili bu açıklamalar incelendiğinde; 28 Temmuz açıklamasında yoğun bakımdaki hastaların çoğunluğunun aşı olmayanlar olması hakkında istatistikler verilirken 11-19 Ağustos açıklamalarında aşı olmayanlar için gelecek kısıtlama ve yasaklar açıklanmaktadır. Resmi kurumların ve/veya devlet yetkililerinin yaptığı açıklamaların, toplumun aşıya karşı tavrına etkisinin büyük olduğu sonucu çıkarılabilir. Bu açıklamalardan bilimsel veriler ile aşının faydalarını anlatan açıklama oldukça pozitif bir etki ile karşılanırken bu konuda yasaklar koymak toplumda bazı şüpheler oluşmasında etkili olduğu yorumu yapılabilir. Sivil toplum kuruluşlarının ve devletin karar verici kuruluşların Covid-19 harici sorunlarda da halkın tepkisini ölçmek için doğal dil işleme ve sosyal medyayı kullanması daha iyi karar almalarında yardımcı olabilir.

## Kaynakça

[1] Q. G. To, K. G. To, V.-A. N. Huynh, N. T. Q. Nguyen, D. T. N. Ngo, S. J. Alley, A. N. Q. Tran, A. N. P. Tran, N. T. T. Pham, T. X. Bui ve C. Vandelanotte, «Applying Machine Learning to Identify Anti-Vaccination Tweets during the COVID-19 Pandemic,» *Int J Environ Res Public Health,* no. 18, p. 4069, 2021.

[2] J. Ruiz, J. D. Featherstone ve G. A. Barnett, «Identifying Vaccine Hesitant Communities on Twitter and their Geolocations: A Network Approach,» %1 içinde *HICSS 2021*, 2021.

[3] B. AYAN, B. KUYUMCU ve B. CİYLAN, «Detection of Islamophobic Tweets on Twitter Using Sentiment Analysis,» *GU J Sci,* cilt 7, no. 2, pp. 495-502, 019.

**For citation and final version please click here:** Küçüktaş,Ü.T. et al. (2021). BERT Modeli ile COVID-19 Aşısı için Aşı Karşıtı Tweetlerin Arttığı Zaman Aralıklarının Tespiti. 1st International Congress on Artificial Intelligence and Data Science (s. 19-24). İzmir Republic of Turkiye: İzmir Kâtip Celebi University.

[4]  J. Xue, J. Chen, R. Hu, C. Chen, C. Zheng, Y. Su ve T. Zhu, «Twitter Discussions and Emotions About the COVID-19 Pandemic: Machine Learning Approach,» *J Med Internet Res.,* no. 11, pp. 25-22, 2020.

[5]  H. Budhwani ve R. Sun, «Creating COVID-19 Stigma by Referencing the Novel Coronavirus as the "Chinese virus" on Twitter: Quantitative Analysis of Social Media Data,» *J Med Internet Res.,* no. 5, pp. 6-22, 2020.

[6]  JustAnotherArchivist, «snscrape: A social networking service scraper in Python,» [Çevrimiçi]. Available: https://github.com/JustAnotherArchivist/snscrape. [Erişildi: 26 08 2021].

[7]  MDZ Digital Library team, «dbmdz Turkish BERT model,» MDZ Digital Library team (dbmdz) at the Bavarian State, [Çevrimiçi]. Available: https://huggingface.co/dbmdz/bert-base-turkish-cased. [Erişildi: 26 08 2021].

[8]  J. Devlin, M.-W. Chang, K. Lee ve K. Toutanova, «BERT: Pre-training of Deep Bidirectional Transformers for Language Understanding,» *arXiv preprint,* 2018.

[9]  Twitter, «28 Temmuz 2021 - Sağlık Bakanı Fahrettin Koca'nın Bilim Kurulu sonrası açıklaması,» [Çevrimiçi]. Available: https://twitter.com/drfahrettinkoca/status/1420455185357361154. [Erişildi: 26 08 2021].

[10]  «Aşı olmayanlar şehirler arası seyahat edebilecek mi?,» TRT HABER, [Çevrimiçi]. Available: https://www.trthaber.com/haber/guncel/asi-olmayanlar-sehirler-arasi-seyahat-edebilecek-mi-asi-olmayanlara-pcr-testi-zorunlulugu-603441.html. [Erişildi: 26 08 2021].

[11]  TRT HABER, «Cumhurbaşkanı Erdoğan: Aşı olmayandan PCR testi isteyeceğiz,» [Çevrimiçi]. Available: https://www.trthaber.com/haber/gundem/cumhurbaskani-erdogan-asi-olmayandan-pcr-testi-isteyecegiz-603030.html. [Erişildi: 26 08 2021].